**JSCDSS**
An International Journal
http://www.jscdss.com
Vol.3 No.4 August 2016: 7-18
Article history:
Accepted 14 May 2016
Published online 14 May 2016

**Journal of Soft Computing and Decision Support Systems**

Journal of Soft Computing and Decision Support Systems

**JSCDSS**

E-ISSN: 2289-8603

# Fuzzy Based Implicit Sentiment Analysis on Quantitative Sentences

Amir Hossein Yazdavar [a,*], Monireh Ebrahimi [a], Naomie Salim [a]

[a] Faculty of Computing, Universiti Technologi Malaysia, Johor, Malaysia

**\* Corresponding author email address**: yazdavar@gmail.com

**Abstract**

With the rapid growth of social media on the web, emotional polarity computation has become a flourishing frontier in the text mining community. However, it is challenging to understand the latest trends and summarize the state or general opinions about products due to the big diversity and size of social media data and this creates the need of automated and real time opinion extraction and mining. On the other hand, the bulk of current research has been devoted to study the subjective sentences which contain opinion keywords and limited work has been reported for objective statements that imply sentiment. In this paper, fuzzy based knowledge engineering model has been developed for sentiment classification of special group of such sentences including the change or deviation from desired range or value. Drug reviews are the rich source of such statements. Therefore, in this research, some experiments were carried out on patient's reviews on several different cholesterol lowering drugs to determine their sentiment polarity. The main conclusion through this study is, in order to increase the accuracy level of existing drug opinion mining systems, objective sentences which imply opinion should be taken into account. Our experimental results demonstrate that our proposed model obtains over 72% F1 value.

## 1. Introduction

Long time before web used as a media for transferring information and opinion, people usually ask each other or their friends to make decision through the various issues such as buying product or planning to vote. However nowadays, there is no limitation to ask others opinion since internet and web provide a vast pool of reviews from millions of people that we even did not know them. According to one survey cited by (Pang and Lee, 2008) more than 80 percent of web users have done at least one online search while purchasing a product. However, monitoring and finding the other's idea might be confusing and overwhelming since finding relevant sites and reliable opinion through a huge volume of opinionated text in each sites, seems to be impossible. Thus, there is a clear need to build an automatic system to help finding and extracting opinions about different topics.

According to (Liu, 2015) opinion can be defined as a quintuple ($e_i$, $a_{ij}$, $s_{ijkl}$, $h_k$, $t_l$) where $e_i$ is the name of entity, $a_{ij}$ denotes an aspect of entity $e_i$, $s_{ijkl}$ shows the sentiment orientation on aspect $a_{ij}$ of entity $e_i$, $h_k$ is opinion holder (the one who express this idea) and $t_l$ is the time when this idea expressed by its opinion holder. By this definition, sentiment analysis is the task to find all quintuples in the given opinionated document. Opinions fall within two categories based on the way they are expressed; explicit opinion and implicit opinion. An explicit opinion includes subjective sentences which have opinion keywords. These kinds of statement are easy to detect. The bulk of current research is devoted to this category. On the other hand, implicit opinion is implied in objective statements. Indeed, they state one desirable or undesirable fact (Kim and Hovy, 2004). Limited work has been done in this category. One special kind of these sentences which is noun and noun phrase that imply opinion has been taken into consideration by (Zhang and Liu, 2011). However, they give a general approach for all domains whereas opinion mining (especially implicit opinion mining) is a highly domain and context dependent task (Rushdi et al., 2011) and to reach good and applicable results it would be beneficial to exploit some domain knowledge and features. One group of such domain dependent problems is related to change or deviation from desired range or value. In some application domains, the values of an item have specific properties which denote the sentiment. Indeed, change of these values to the optimal interval or deviation from the norm range might express positive or negative opinion respectively. For example, let the optimal value for total cholesterol have been defined below than 200 mg/dl, then the sentence "this drug lowered my cholesterol form 300 to 190" implies positive sentiment since the cholesterol decreased into optimal value. And in sentence "don't take this drug, it puts my Blood pressure into 18" implicit opinion by deviating from normal range has been expressed. Furthermore, based on our observation, significant changes might also denote sentiment even the new value would not be placed in a normal range. For instance in the sentence "it dropped my cholesterol level from 580 into 250", although the second value of cholesterol would not be placed in the optimal value (below than 200mg/dl) but the sentence express positive sentiment. On the other hand, the sentence "it increased my cholesterol level from 250 to 580" denotes negative sentiment polarity, thus it is important to consider change direction. It is worth mentioning to recall that, among these kinds of sentences, there exist some kinds of sentences that do not express any sentiment while containing changes in numeric values. As an example "my doctor changes the normal dosage of Welchol from 624mg to 300mg" is a factual sentence that should be grouped into non-opinionated sentence.

In this regard, one special characteristic of such sentences is related to certain degree of uncertainty and imprecision involved in them. In the light of our observation, this uncertainty can be regarded from different points of view. First, a large portion of quantitative medical terms associated with predefined ranges which try to identify a patient's status e.g. for total cholesterol, below than 200mg/dl considered as desirable, between 200 and 239 is borderline high and greater than 240 is defined as high. Thus, these terms can be accepted as fuzzy. Second, changes in the value of these quantitative terms, might demonstrate improvement, stable condition, and exacerbation of a patient, regardless of where the second value is placed. For example, the sentence "my total cholesterol dropped from 580 to 240" shows improvement in total cholesterol while the second value of it (240) is placed into high range. Indeed fuzzy set theory is an excellent approach to deal with inexact medical entities for approximating medical text (Hatiboglu et al., 2010; Keleş et al., 2011; Samuel et al., 2013). Intuitively, changes can be grouped into slight, medium and high increase or decrease which can be denoted by fuzzy set theory. Furthermore, patient's sentiment might be regarded as positive, neutral and negative based on the factors that have been stated including second numeric value and changes and also side effect, opinion words and etc. Consequently, fuzzy logic is an ideal choice to deal with this problem.

In the present paper, we investigate special kind of numerated sentences, not only to categorize them into opinionated and non-opinionated but also to determine whether they contain positive or negative sentiment polarity by employing fuzzy set theory. These kinds of factual sentences contain quantitative measurement term that implicitly express sentiment. Indeed, the use of numerated objective sentences to determine sentiment polarity by adapting fuzzy knowledge based engineering techniques is the main contribution of our study.

There are four main issues which can achieve the goal of this study; what is the method to extract numeric variable in a sentence? How this numbers are related to their specific entities? Which numerated sentence bear a sentiment? And what is the sentiment orientation of that opinionated sentence?

The paper is organized as follows. Next section comments on some related work and approaches in sentiment analysis particularly in biomedical domain. Some studies on application of fuzzy in this domain are also covered. A detailed definition of problem which is solved in this study is given in section 3. Section 4 presents the method applied and the experiments carried out. Results obtained are reported and discussed in section 5 and 6. Finally, in section 7 the main conclusion and proposals for future work are expounded.

## 2. Related Works

Sentiment words can play an important role in sentiment classification task. In this regard, (Turney, 2002) apply an unsupervised method to perform classification. His suggested algorithm consists of; first, defining some syntactic pattern based on part-of-speech (POS) tags and then extracting two consecutive words based on these patterns. Second, using pointwise mutual information (PMI) to estimate the sentiment orientation of extracted phrase. Although this simple idea performed well in some areas, subsequent researches have been improved this method. (Yu and Hatzivassiloglou, 2003) used the similar method to (Turney, 2002) but instead of using PMI, they applied log-likelihood ratio. Furthermore, they employed a large set of adjective seeds instead of one word for positive and one word for negative. Another related work based on bootstrapping strategy and employing sentiment lexicon proposed by (Hu and Liu, 2004). Their algorithm starts with some given positive and negative sentiment words as a seed. To generate lexicon, they used WordNet to find synonyms and antonyms relation. For each positive word they considered +1 and -1 for negative word. Summing all the sentiment word's orientation, determine sentiment of each sentence. Determining sentiment orientation on each aspect in a sentence defined as aspect sentiment classification. In recent years there has been a wealth of research in methods for aspect extraction (Chen, Mukherjee, and Liu, 2014) (Chen et al., 2013). In particular, to cope with this problem two main approaches namely supervised learning approach and lexicon based approach have been studied extensively (Rana and Cheah, 2016). Although supervised manner have been exploited widely by researchers a problem inherent in this method is its dependency to training data. Thus, it is difficult to extend this approach to various domains. On the other hand, lexicon based approach performs quite well in a large number of domain. One basic method, the foundation of most other later works, has been proposed by (Ding et al., 2008). The main idea behind their method was sentiment words surrounding each product aspect in sentence can be denoted the sentiment orientation on the product aspect.

Despite the importance of analyzing sentiment in medical domain, limited research has been done. In particular social media data is a huge source of experiences on drugs, treatments, diagnosis, drug side effects (Denecke, 2015). This experiential knowledge can be utilized as an extension to the facts in healthcare treatment.

(Yalamanchi, 2011) proposed a negativity meter system considering drug side effect. They have argued, unlike ordinary sentiment analyzing product reviews, considering sentiment words and subjective sentence would not be effective in drug reviews since there exists a large number of objective sentences which imply sentiment. For instance in review "Here are the side effects I experienced: severe depression, suicidal ideation, attempted suicide, mania, impulsiveness and impaired thinking" although there is not any subjective word, the reviewer expresses the negative sentiment about the drug implicitly. Thus, side effects need to be taken into consideration for determining sentiment polarity in such domain. In other related work, (Swaminathan et al., 2010) studied relationship between bio-entities and defined new features for SVM classifier and combined them with lexicon based approach to predict polarity. They also identified the strength of relationship using SVR. Another challenge associated with drug reviews is mentioned by (Liu and Zhang, 2012). They suggested considering the time when expressed desirable or undesirable fact occur. Indeed, these kinds of sentence only imply opinion when occur after taking the drug not before. For example the sentence "when my doctor gave me Methadone, I had headache for ten years" does not show any sentiment because "headache" occur before utilizing this drug.

Besides, extracting quantitative medical entities such as LDL and HDL play a pivotal role in biomedical opinion mining. (Bhatia, et al., 2010) extracted numeric fields utilizing regular expressions. Their algorithm loops through all of the token and testing numeric token against a set of regular expressions. In other related work, (Rosier et al., 2008)suggested the method based on General Architecture for Text Engineering (GATE) to create rules based on syntactic and semantic co-occurrence patterns found on domain knowledge.

Furthermore, there are number of methods which have been employed to determine relationship among pair of entities. In this regard, a large amount of researches have suggested entities co-occurrence which provides high recall and low precision in accuracy (Cao et al., 2007). Unlike this method, rule based approach that describe linguistic patterns, provides high precision and low recall (Abacha and Zweigenbaum, 2011). Another approach to deal with this problem is machine learning methods.(Simpson and Demner-Fushman, 2012)consider syntactic structures causes to introduce another method which applied dependency parser that is capable to encode grammatical relations between phrases and words. (Zhou et al., 2006) applied link grammar parser to map numeric attributes to its corresponding medical concept for determining patient with breast cancer. In another related work, (Zhou et al., 2005) combined the link grammar parser method and some linguistic patterns to convert semi structural medical note to structured text. In the present study, we will adapt linguistic patterns and heuristics approach to associate extracted aspect and number in a sentence. The main advantage regarding this approach is its simplicity and better performance and high precision.

On the other hand, fuzzy set theory has been studied by many researchers in fields that uncertainty plays an integral role. Since, medicine contains a large amount of uncertain information such as measurement imprecision, vagueness, hesitation and natural diversity subjectivity this domain can be regarded as uncertain domain. As an example, consider the word "high" for blood pressure; from the expert point of view, there is not a precise definition value for high blood pressure or for hypotonic and hypertonic patient high blood pressure may be defined quite different. Additionally, different situations require flexible interval determining of these measurements for instance in anaesthesia which range refers to high? To deal with this large amount of uncertain knowledge, many fuzzy based systems have been proposed. Abbodet al. (2001) presented a comprehensive survey of fuzzy application in medical domain.

Another major issue which has been unexplored to date is identifying sentiment orientation of sentences which contain number in drug reviews. In the present paper, we will try to remedy this problem using fuzzy method.

## 3. Problem definition

A large number of patients in order to express the opinion toward the effectiveness or side effect of drugs that they took, use some quantitative values. For example, sentence "Using this drug, my cholesterol level went from 518 to 175" implies positive sentiment about this drug, or sentence "Taking this drug, my blood pressure rise to 18" demonstrates negative sentiment on this drug since this drug causes "blood pressure" to deviate from its normal range. Based on our observation, objective sentences which imply sentiment can be divided into 3 main categories. The first group includes sentences which contain 2 values of one specific medical term as an example sentence "NIASPAN, brought my LDL from 175 to 105" contains 175 and 105 for the medical term LDL. The second group is dedicated to sentences which denote the change of a specific medical term. For example sentence "This lowered my "bad" cholesterol 25 pts in two months" show the change in cholesterol by 25 point or in sentence "First three months, HDL rose 11 points" HDL has been improved by specified drug. Changes expressed in percent in some cases like "Worked great to reduce my cholesterol by about 20% within just a few weeks". And finally, sentences contain 1 value of specific medical term and special kind of verb such as "lowered", "brought", "fell", "changed", "decline", "dropped", "rise", "shot", "increased", "decreased", "down", "up", "improve" and etc. For instance sentence such as "Cholesterol fell to 160" or "This drug increased my HDL to 55", has potential to consider as an opinionated objective sentence. Nevertheless, there exist a large amount of numerated sentences in drug reviews that do not show any sentiment and state some facts. For example sentence "My doctor want my cholesterol went down to 150" although contains "down to" as a verb, does not show any sentiment. Table 1 further illustrates the point. We notice from Table 1 opinionated sentences typically contain special aspects such as blood pressure, heart rate, blood sugar, pulse, period cycle, cholesterol, HDL, LDL, triglycerides. On the other hand, numbers denoting; dosages, number of time a drug has been taken and how long a patient suffer from his/her diseases, state fact and need to be filtered. In addition, there are some aspects that require special treatment, for example the sentence "I lost 13 lbs on this drug without effort (from 127 to 114)" implies sentiment on weight aspect whereas in sentence "My weight is around 128 - 130" no sentiment has been expressed on this aspect. Thus, a proper methodology is required to discriminate opinionated sentences from factual statements and determine the sentiment polarity of opinionated one.

Hence, in this paper we study the objective quantitative terms which implicitly express sentiment. Based on our observations, biomedical texts are the rich set of numeric fields and their values. Indeed, health monitoring accomplished by measuring quantitative terms which their values denote the level of health. "Blood pressure", "weight", "glucose fasting", "total cholesterol" including LDL and HDL cholesterol and "triglyceride" are common entities of this type. Further details on proposed approach to deal with this problem are given in the next section.

## 4. Proposed Approach

### 4.1 Overview

In this research, some reviews were collected from patient's review for building corpus. Medical experts cooperated in this research for building gold standard corpus. Pre-processing operations including extracting medical terms and their corresponding values have been done on this corpus. An appropriate technique developed to map each of these medical terms to their corresponding values. Determining sentiment polarity of each sentence employing fuzzy knowledge based system is the next study of this research. Finally, precision and recall were applied to evaluate the result of suggested algorithm. Fig. 1 illustrates the summary of whole process.

### 4.2 Collecting Drug Reviews

In this study, 210 drug reviews were collected from website www.askapatient.com. These reviews are selected from different cholesterol lowering drugs. One group of cholesterol lowering drugs named Statin. Statins contain various types that each of them provides different characteristics. There are five main categories of Statins namely; Lovastatin, Pravastatin, Simvastatin, Atorvastatin and Rosuvastatin. Taking each of them, affect patients differently (Jones, et al., 2003).Moreover, another therapy to lower cholesterol is nicotinic acid (niacin). The effect of this type has been studied extensively in (Capuzzi, et al., 1998). We collect our sentences from aforementioned categories of cholesterol lowering drugs. However in order to provide an efficient expert system, different aspects of these drugs have been taken into consideration by the aid of medical experts.

Some of the characteristics of constructed corpus have been presented in Tables 2 and 3. Table 2 represents the number of reviews and sentences which have been extracted from web to build gold standard corpus. It also represents the polarity distribution of all sentences exists in the corpus. Constructed corpus contains four types of numerated opinionated sentences. Table 3 illustrates the distribution of these sentences with their sentiment.

**Table 1**
Different types of numerated sentences in medical reviews.

| | Sentence Type | Numeric Fields | Sample sentences |
|---|---|---|---|
| **Potentially opinionated** | From *First value* To *Second value* | Cholesterol | My cholesterol went from 279 to 230. |
| | | LDL | LDL Went from 201 to 65. |
| | | HDL | Increased HDL from 51 to 86. |
| | | Triglycerides | Niaspan dropped my Triglycerides level from 311 to 175 over a one month period. |
| | Change by *Percent* | Cholesterol | The drug did knock my Cholesterol down by 55%, so I have motivation to try and make it work. |
| | | LDL | Improve my LDL 30 %. |
| | | HDL | Lpa from 33 to 5 HDL up 20%. |
| | | Cholesterol, HDL, LDL, Triglycerides | Reduced Total 27%, Trig 40%, LDL 32% and increased HDL 17%. |
| | Change by *Amount* | Triglycerides | Good stuff, just 500 mg lowered my trig. 30 points in a month. |
| | | Cholesterol | This dropped my Cholesterol by 30 pts w/in a couple of months. |
| | | Cholesterol, HDL, LDL | My cholesterol dropped 5 points, my HDL dropped 3 points, and my LDL raised 1 point. |
| | Changed into *Final value* | Cholesterol | After 3 months, I lowered my Cholesterol total to 167. |
| | | HDL | After 90 days of 500mg, HDL raised to 34, not good enough. |
| | | LDL | By the way 10 mg dropped my LDL to 77. |
| **Non-opinionated** | Facts | Consuming time | I took at 9:30 pm and it is now 8 am |
| | | Dosage | So I decided to cut my pills in half and go back to 25mg. |
| | | Age | I am only 36 and my husband is a young 40 |
| | | Duration time | Drug is helping hair loss, was told it would take from 6 to 12 months to start to work. |

**Table 2**
Dataset statistics

|  | Number |
|---|---|
| Reviews | 210 |
| Sentences | 2344 |
| Tokens | 19160 |
| Numbers | 1991 |
| Positive Sentences | 165 |
| Neutral Sentences | 45 |
| Negative Sentences | 25 |

**Table 3**
Polarity distributions over different types of sentences.

|  | Change from 1st Value to 2nd Value | | Changed into Final Value | | Change by Amount | | Change by Percent | |
|---|---|---|---|---|---|---|---|---|
|  | n | % | n | % | n | % | n | % |
| Positive | 128 | 55 | 22 | 10 | 8 | 3 | 6 | 3 |
| Neutral | 28 | 12 | 9 | 4 | 6 | 3 | 2 | 1 |
| Negative | 8 | 3 | 8 | 3 | 5 | 2 | 1 | 1 |

*4.3 Preprocessing*

The reviews need to be processed employing some NLP external tools. In this study, we utilize GATE not only extract numbers and medical terms but also for mapping the extracted medical terms into their corresponding values by applying Java Annotation Patterns Engine (JAPE) Grammar. GATE is one of the most common software tools in processing human language. Additionally, GATE has been distributed with an information extraction system named A Nearly-New Information Extraction (ANNIE). ANNIE includes different components (Cunningham, 2002). We employ these components in order to accomplish various tasks as illustrated in Fig. 2. As the figure depicts after tokenization and sentence splitting, Gazetteer has been used to extract medical quantitative entities such as LDL and HDL cholesterol based on some predefined lists. Indeed, for extracting each medical term and numeric values, we prepare a list contains various form of each medical term and values which people employed in their comment to express their opinion. Table 4 represents some sample of such lists. Furthermore, Gazetteer has been exploited for determining the verb which shown a change in the text. In this regard, two sets of such word has been compiled which one of them state "increase" whereas the other one express "decrease".

On the other hand, one of the major challenges through this study is to associate the extracted numbers with recognized medical terms due to existence of more than one target pair e.g. "My LDL is 64, HDL of 74 and TC 152". In this study, we will apply a rule based approach using regular expressions for extracting relations between numbers and medical terms. At the end of this phase, each medical term and its corresponding value and also drug name and its dosage have been added as a feature to every sentence. Our tailored method utilizes Jape Transducer which is based on regular expression. Simplicity is one of the major advantages of the proposed method. Another benefit regarding the suggested method is its ability for handling verb less sentences such as "HDL: 175". Some of these JAPE rule patterns are demonstrated below.

**Table 4**
Various forms of quantitative medical terms employed by patients

| Cholesterol | LDL | HDL | Triglyceride |
|---|---|---|---|
| Chol. | bad cholesterol | good cholesterol | TG |
| CHOL | bad choles | good cholesterol | Tri |
| total C | bad chol | good cho | Trig |
| Total-C | bad cho | good chol | Trigl. |
| total choles | Bad-C | HDLS | triacylglyceride |
| Total Col | bad cholesterol | good C | Tryglycerides |

*4.4 Fuzzy Logic Approach*

Fuzzy modeling which is introduced by (Zadeh, 1965) is a powerful approach for modeling of complex and uncertain systems. Indeed, fuzzy logic has a particular advantage in fields were precise definition of control process is unachievable. Thus, medical decision making is one of well-suited area for applying fuzzy logic. Fuzzy models have ability to establish relationship between input and output variables by employing predefined rules. This methodology obtains simple solution rather than statistical issues which is based on natural language statements (Hatiboglu et al., 2010). The basis of fuzzy logic is to consider the inputs and outputs in the form of fuzzy sets. Each fuzzy set contains of elements that have varying degrees of membership. Fuzzy set enables to transform real number to the membership degrees changing from 0 to 1(Ross, 2009). Fuzzy rules relate input variables to output variables. These rules represent expert's knowledge in the system. Mamdani's approach is one of the most common fuzzy inference methods which defines the output of each fuzzy rule as a fuzzy set for the output variable (Mamdani, 1974). Defuzzification is the last step over the system which is a mapping process from a fuzzy space defined of output into crisp.

In this paper, we apply fuzzy set theory to determine sentiment orientation in drug reviews. To address this issue, the proposed method consists of following stages; first, determining inputs and outputs variable. Then, specify fuzzy set for these inputs and outputs. Next, design membership function of all variables. Then, ask experts to generate fuzzy rules based on these variables. Finally, employ an inference approach to accomplish defuzzification process. More details will be given in next section.

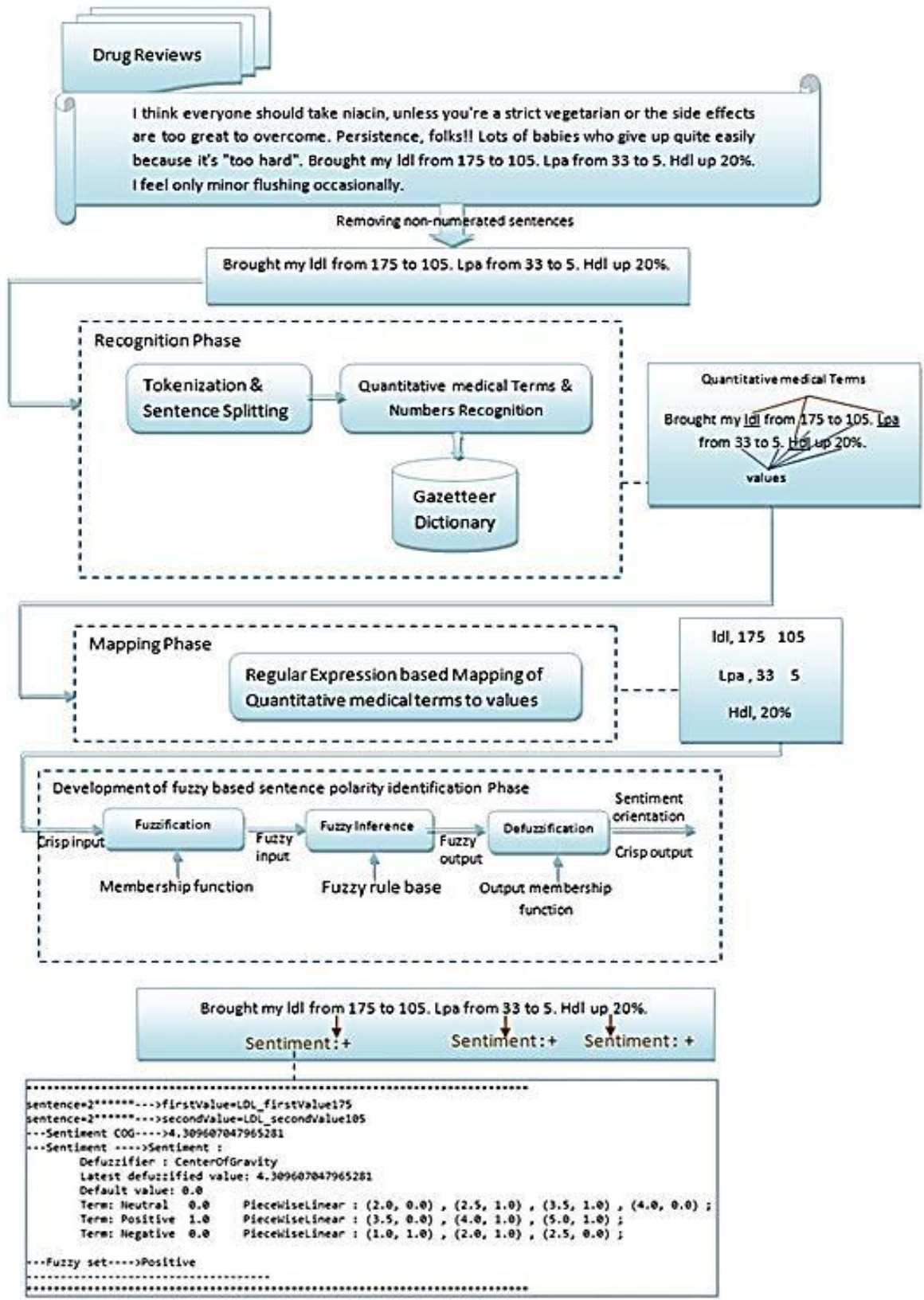

**Fig. 1.** Diagrammatic summary of proposed approach

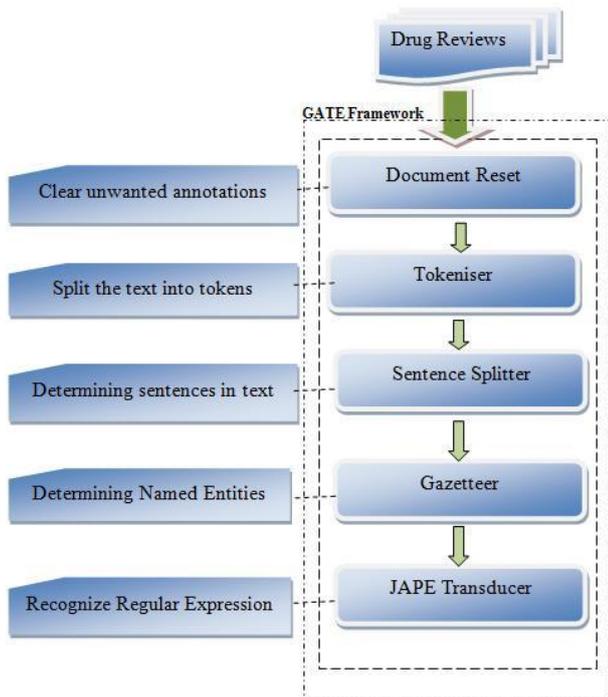

**Fig. 2.** Gate pre-processing framework

*4.5 Development of Fuzzy Logic Technique*

After extracting the interested sentences with their features, determining the sentiment polarity of each sentence is the next step. At the end of this phase our system is able to identify whether a sentence bear positive, negative or neutral polarity toward drug effectiveness. Our suggested system consists of first, knowledge acquisition, second, design of fuzzy expert system exploiting fuzzy rule based system, and finally, implementation of the entire system.

*4.5.1 Knowledge Acquisition by Interview and Literature*

Knowledge acquisition (KA) is one of the most time-consuming and complicated part for developing an expert system. The main aim of KA in our method is to efficiently acquire specific knowledge from a group of medical experts for determining the effectiveness of a specific drug. Multiple interviews with medical expert from different medical centers have been done. The most important factors responsible for determining whether a specific drug was effective for a patient or not, are identified. Below, we briefly explain some of our findings through interviews as well as literature.

As mentioned earlier, one of the main class of cholesterol lowering drugs are Statins. Indeed, Statins are most commonly prescribed drug to decrease LDL cholesterol. Besides, Statins play an integral role for cardiovascular diseases (CVD) treatment. According to the medical experts, there are 5 types of Statins. (Jones, et al., 2003) compare these types by variant dosage on 6 week trial. According to their investigation, the dosage and type of each cholesterol lowering drug might causes variety of change from base line. They also provide Mean Percent Change from base line (BL) in HDL, Triglyceride and Total cholesterol. Based on these observations, drug type and dosage are two important criteria which should be taken into consideration in order to design an efficient expert system.

Niacin is another category of cholesterol treatment drugs specifically for increasing HDL cholesterol. Based on the medical literatures, they are able to increase HDL between 15 and 35 percent. (Capuzzi, et al., 1998) study the effect of Niacin (Niaspan) on patients with primary hypercholesterolemia. Their method includes two parts, firstly, every patient was given a 4 week titration pack of Niaspan and their dosage was initiated at 375mg of this drug. After these 4 weeks, Niaspan dosage prescribed for 1000-3000 mg depending on patient therapeutic response and evaluation of adverse experience. Their study shows the significant effect of duration of taking Niaspan for increasing HDL cholesterol.

Based on our findings, not every cholesterol lowering drugs works for every type of hypercholesterolemia. Indeed, some are more effective for increasing HDL whereas some are better for LDL. Additionally, it is doctor's decision to prescribe each type of these drugs and their dosage based on patient's condition.

To sum up, in the present paper, to analyze whether a drug was effective or not, type of each drug, dosage and duration, has been considered by medical experts who aid us to design our expert system.

*4.5.2 Structure of Proposed Fuzzy Expert System*

Our expert system contains three major components; knowledge base (KB), Inference engine and Input data. KB includes the medical expert information and knowledge which is encoded in computer perceptible form. Inference engine contains the ability of reasoning to decide which rules need to be fired. The input parameters should be fuzzified which means each element has a partial belongingness to one or more than one sets. This partial belongingness denoted by membership function.

Recall from previous section, we determine the fuzzy variable with the aid of medical expert. Each fuzzy variable has different fuzzy sets. Linguistic representation of some variables has been shown in Table 5. Each fuzzy variable has been represented by fuzzy membership function. We exploited trapezoid membership function since it contains more information of fuzzy sets. Indeed, the normal and abnormal ranges for every input have been extracted by literatures as well as doctor's opinions. The output variable of proposed expert system denotes the "Sentiment" of a patient after taking a drug on the specific sentence. Therefore, "sentiment" defines as a fuzzy variable with the fuzzy sets of negative, neutral and positive. Some samples of input and output membership graphs have been plotted in Figs. 3 and 4 respectively.

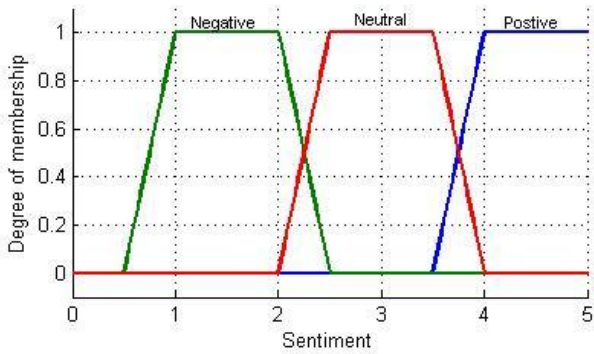

**Fig. 3.** Sentiment fuzzy set

Fuzzy knowledge base has been developed with the aid of doctor's past experience. After multiple rounds of interviews with medical expert, the preliminary rules have been extracted. These rules were verified by the other doctor. However, there were some conflicts which have been resolved by the consensus of medical experts. Table 6 shows some sample of generated fuzzy rules. Our designed system adopts Mamdani's inference approach. Finally, for defuzzification step "Centroid" method has been utilized.

## 5. Results

In this study, precision and recall and F-measure were applied to evaluate the result of suggested algorithm. Due to novel nature of this problem there is not any annotated corpus for testing and evaluation of the result of this study. Thus, human experts participated in this work by determining sentiment orientation of all sentences within a corpus and building a gold standard corpus. These assigned sentences' polarities have been used to calculate precision and recall.

Evaluation task falls into following sub tasks; first, evaluate the accuracy of values and medical terms extraction. Second, examining the efficiency of Jape Grammar method proposed for determining each medical term-value pair. Finally, evaluate sentiment classification output with assigned sentiment in the reference data set. Overall precision and recall of the proposed system is shown in Table 7.

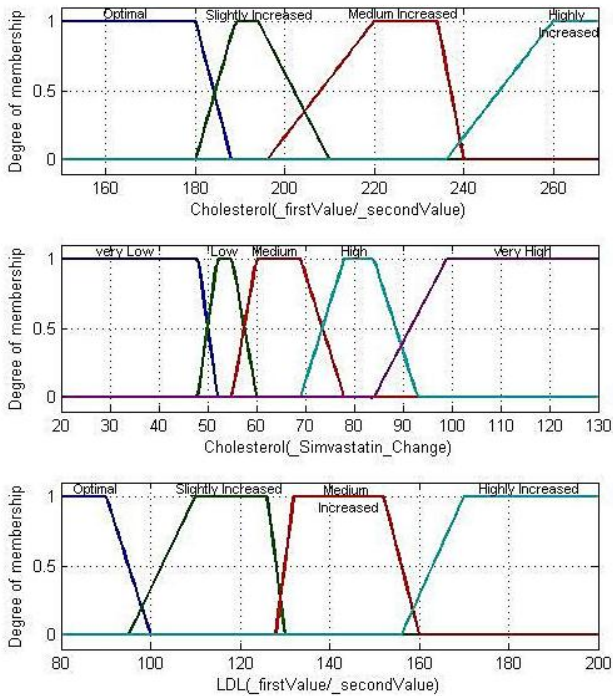

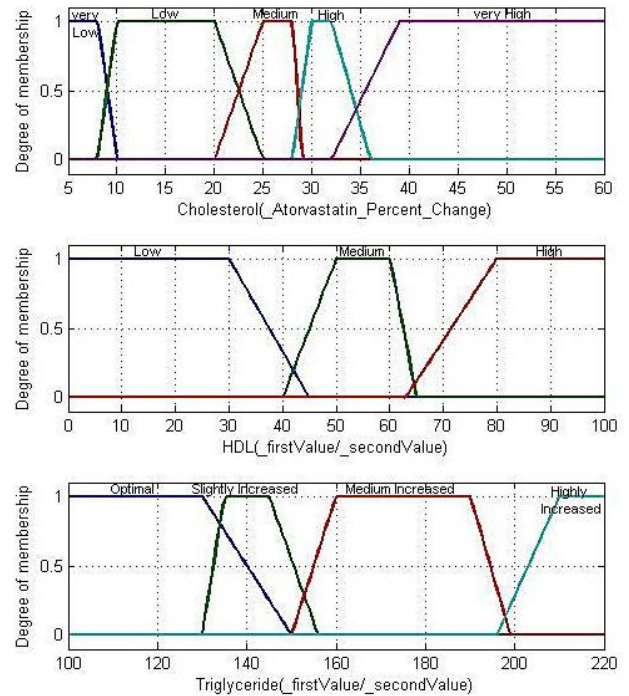

**Fig. 4.** Some samples of Input fuzzy sets

Furthermore, in order to do error analysis more precisely the confusion matrix has been shown in Table 8 and some of false positive and false negative results are examined. In this regard, following issues are the most salient part of our findings. For some sentences, changing of the medical term is not the effect of taking its corresponding drug. As an example, in sentence "After dropping 24 lbs & with regular exercise, my total cholesterol went down to 124 (I have lost a total of 70 lbs over 3 yrs)", although the cholesterol down to its desirable amount, but it was because of exercising. To mitigate this problem, extraction of subject exploiting available parser might be effective. Besides, "time" is one of the important factors for determining the sentiment orientation in medical reviews. For instance, in sentence "HDL's went from 24 to 39 in just under 3 month" although there is slight change in HDL, but it was in a short period of time thus, it must be considered as a positive sentence. Temporal (Solution). Furthermore, there are some sort of sentences which considered as opinionated sentences by our algorithm while they state some facts. As an example sentence "My Doctor wants my chols went down to 150". Conditional sentences are the other types of sentences which require special treatment. As an example, sentence "The cardiologist agreed to go to Lipitor if they could drive my number below than 100" state some facts. Moreover, co-reference resolution is one of the major challenges in every NLP problem which has not been solved completely yet. A large number of sentences have been affected by this problem. For example in sentence "It lowered my total cholesterol to 99. But being it was only 170 when I started treatment that is not a huge accomplishment" the second "it" denotes cholesterol. In addition to all of these challenges, our extraction algorithm also confronts many issues through this study. Verb less sentences is one of these issues. Although our extraction algorithm tries to deal with verb less sentences, there are large numbers of cases which it is not comprehensive enough to extract their entities properly. As an example sentence like "approx results-3 month intervals dose/hdl/ldl/TriG: 500/29/87/236; 500/23/86/225; 1000/28/77/246; 1500/22/69/182. Not overwhelming" requires special pattern for extracting and mapping each of the medical term to their corresponding values.

There are other types of numerated opinionated sentences that our method was not sophisticated enough to extract them. Table 9 indicates some sample of these sentences.

**Table 5**

Sample of used Fuzzy sets

| Linguistic Variables | Fuzzy Sets |
|---|---|
| Duration | Complete, incomplete |
| Drug Dosage | Low, Medium, High, very High |
| Total Cholesterol | Optimal, slightly increased, medium increased, Highly increased |
| LDL | Optimal, slightly increased, medium increased, Highly increased |
| Triglyceride | Optimal, slightly increased, medium increased, Highly increased |
| Cholesterol Simvastatin Change | Very low, low , medium, High, very High |
| Triglyceride Simvastatin Change | Very low, low , medium, High, very High |
| LDL Simvastatin Change | Very low, low , medium, High, very High |
| HDL Simvastatin Change | low , medium, High |
| Cholesterol Simvastatin Percent Change | Very low, low , medium, High, very High |
| Triglyceride Simvastatin Percent Change | Very low, low , medium, High, very High |
| LDL Simvastatin Percent Change | Very low, low , medium, High, very High |
| HDL Simvastatin Percent Change | low , medium, High |

## 6. Discussion

This study provides new insights into the sentiment recognition in opinion mining problem. Document corpus was built and annotated by medical experts to develop and evaluate the proposed technique namely fuzzy based knowledge engineering. Based on literatures, developing a system for determining sentiment polarity from the change of variables, was unexplored to date.

The findings indicate that employing fuzzy set theory to deal with medical vagueness in order to determine sentiment of medical sentences seem to be promising way forward in medical sentiment analysis. However, from the result it becomes apparent that the proposed fuzzy system is unable to correctly determine sentiment polarity of some statements. To mitigate this problem, generated fuzzy rules need to be expanded by consensus of more medical experts in order to provide more reliable system. Besides, fuzzy knowledge based system can be easily combined with other machine learning algorithm such as Neural Network in order to circumvent the need for generating fuzzy rules which was one of the time consuming tasks in this study.

Besides, there are some groups of challenging sentences for polarity identification part of our proposed system. Our algorithm classifies these kinds of sentences into the wrong category due to their complex structure. Conditional sentences are one example of these statements. We should expand our algorithm to make it comprehensive enough to deal with these special cases.

Furthermore, one of the major drawbacks involved in this research was its dependency to extraction algorithm. The main reason behind this phenomenon is all possible scenarios has not been taken into account for developing reliable extraction method and this is due to difficulty of analyzing natural language in the presence of verb less sentences, grammatical mistake and typographical error. To alleviate this problem, employing other extraction approaches like supervised and semi supervised algorithm might be other possible solutions.

Moreover, in order to enhance the applicability and feasibility of suggested approach for extraction, ontology based extraction technique considering intricate relationship between various entities which employs deep parsing techniques can be an ideal choice to develop more reliable system.

**Table 6**
Sample of generated fuzzy rules

| Applicable rule # | 'if' clause | 'then' clause |
|---|---|---|
| 1 | IF CHOLESTEROL_DRUG IS Niacin AND Duration IS Complete AND CHOLESTEROL_Niacin_Change IS Medium | Sentiment IS Neutral |
| 2 | IF CHOLESTEROL_DRUG IS Niacin AND Duration IS Complete AND CHOLESTEROL_Niacin_Change IS High | Sentiment IS Positive |
| 3 | IF CHOLESTEROL_finalValue IS Optimal | Sentiment IS Positive |
| 4 | IF CHOLESTEROL_finalValue IS Highly_Increased | Sentiment IS Negative |
| 5 | IF CHOLESTEROL_finalValue IS Medium_Increased | Sentiment IS Neutral |
| 6 | IF CHOLESTEROL_DRUG IS Simvastatin AND Drug_Dosage IS High AND Duration IS Complete AND CHOLESTEROL_Simvastatin_Percent_Change IS Medium | Sentiment IS Neutral |
| 7 | IF CHOLESTEROL_DRUG IS Atorvastatin AND Drug_Dosage IS High AND Duration IS Complete AND CHOLESTEROL_Atorvastatin_Change IS very_Low | Sentiment IS Negative |
| 8 | IF LDL_DRUG IS Simvastatin AND Duration IS Complete AND Drug_Dosage IS High AND LDL_Simvastatin_Change IS Medium | Sentiment IS Neutral |
| 9 | IF CHOLESTEROL_firstValue IS Highly_Increased AND CHOLESTEROL_secondValue IS Highly_Increased AND CHOLESTEROL_DRUG IS Simvastatin AND Drug_Dosage IS High AND Duration IS Complete AND CHOLESTEROL_Simvastatin_Change IS very High | Sentiment IS Positive |

**Table 7**
Overall precision and recall

| Overall precision | Overall recall | Overall F1 |
|---|---|---|
| 0.81 | 0.64 | 0.72 |

**Table 8**
Confusion matrix for proposed algorithm

| Actual \ Predicted | Neutral | Positive | Negative |
|---|---|---|---|
| Neutral | 30 | 6 | 7 |
| Positive | 19 | 139 | 5 |
| Negative | 7 | 0 | 15 |

**Table 9**
A sample of false positive & false negative result

| No. | Sentence |
|---|---|
| 1 | Three month blood work showed HDL 49-->51, LDL 231-->201, ratio improved from 5.0-->4.1. |
| 2 | My Triglycerides were 871, now down to almost normal range. |
| 3 | Before Niaspan: T-Chol 328, Trig 304, LDL 222, HDL 46. After Niaspan: T-Chol 181, Trig 150, LDL 100, HDL 52. |
| 4 | HDL was 33 and now its 47. |
| 5 | approx results-3 month intervals dose/hdl/ldl/TriG: 500/29/87/236; 500/23/86/225; 1000/28/77/246; 1500/22/69/182. Not overwhelming |
| 6 | LDL 81=>61, HDL 38=>42, total chol 130=>112. I'm taking both Zocor (20mg) and Niaspan. |

## 7.  Conclusion

We developed a novel technique to determine sentiment in the drug reviews based on change of numeric attribute which can be used as a sub task in developing a drug efficiency meter system that uses customer feedback. The main findings through this study is that there exist some type of objective sentences that do not contains any sentiment words whereas they express sentiment and can be regarded as an opinionated sentence. Indeed, in some applications, changing in one item's value causes producing sentiments. Exploiting rule based approach and fuzzy set theory is a reliable method for discriminating these kinds of sentences from non-opinionated one and determining their sentiment. We showed that the proposed approach is very useful in explaining the attitude by the authors toward the efficacy of a target drug.

The result of the proposed algorithms is encouraging and a good foundation for future researches. However, enlarging the corpus which contains various type of quantitative medical term and also consulting with more medical expert annotators in order to construct better gold standard corpus and acquire more medical knowledge and encode it into fuzzy rules would improve the performance of proposed system. Obviating these limitations and exploiting hybrid approaches would improve the results.

## Acknowledgments

This work was partially supported by Fundamental Research Grant Scheme (FRGS) funded by the Malaysia government.

## References

Abacha, A. B., & Zweigenbaum, P. (2011). Automatic extraction of semantic relations between medical entities: a rule based approach. Journal of biomedical semantics, 2(Suppl 5), S4.

Abbod, M. F., von Keyserlingk, D. G., Linkens, D. A., & Mahfouf, M. (2001). Survey of utilisation of fuzzy technology in medicine and healthcare. Fuzzy Sets and Systems, 120(2), 331-349.

Bhatia, R. S., Graystone, A., Davies, R. A., McClinton, S., Morin, J., & Davies, R. F. (2010). Extracting information for generating a diabetes report card from free text in physicians notes. Paper presented at the Proceedings of the NAACL HLT 2010 Second Louhi Workshop on Text and Data Mining of Health Documents.

Cao, H., Hripcsak, G., & Markatou, M. (2007). A statistical methodology for analyzing co-occurrence data from a large sample. Journal of biomedical informatics, 40(3), 343-352.

Capuzzi, D. M., Guyton, J. R., Morgan, J. M., Goldberg, A. C., Kreisberg, R. A., Brusco, O., & Brody, J. (1998). Efficacy and safety of an extended-release niacin (Niaspan): a long-term study. The American journal of cardiology, 82(12), 74U-81U.

Chen, Z., Mukherjee, A., Liu, B., Hsu, M., Castellanos, M., & Ghosh, R. (2013, October). Exploiting Domain Knowledge in Aspect Extraction. InEMNLP (pp. 1655-1667).

Chen, Z., Mukherjee, A., & Liu, B. (2014). Aspect Extraction with Automated Prior Knowledge Learning. In ACL (1) (pp. 347-358).

Cunningham, H. (2002). GATE, a general architecture for text engineering. Computers and the Humanities, 36(2), 223-254.

Denecke, K. (2015). Sentiment Analysis from Medical Texts. In Health Web Science (pp. 83-98). Springer International Publishing.

Ding, X., Liu, B., & Yu, P. S. (2008). A holistic lexicon-based approach to opinion mining. Paper presented at the Proceedings of the 2008 International Conference on Web Search and Data Mining.

Hatiboglu, M. A., Altunkaynak, A., Ozger, M., Iplikcioglu, A. C., Cosar, M., & Turgut, N. (2010). A predictive tool by fuzzy logic for outcome of patients with intracranial aneurysm. Expert Systems with Applications, 37(2), 1043-1049.

Hu, M., & Liu, B. (2004). Mining and summarizing customer reviews. Paper presented at the Proceedings of the tenth ACM SIGKDD international conference on Knowledge discovery and data mining.

Jones, P. H., Davidson, M. H., Stein, E. A., Bays, H. E., McKenney, J. M., Miller, E., . . . Blasetto, J. W. (2003). Comparison of the efficacy and safety of rosuvastatin versus atorvastatin, simvastatin and pravastatin across doses (STELLAR Trial). The American journal of cardiology, 92(2), 152-160.

Keleş, A., Keleş, A., & Yavuz, U. (2011). Expert system based on neuro-fuzzy rules for diagnosis breast cancer. Expert Systems with Applications, 38(5), 5719-5726.

Kim, S.-M., & Hovy, E. (2004). Determining the sentiment of opinions. Paper presented at the Proceedings of the 20th international conference on Computational Linguistics.

Liu, B. (2015). Sentiment analysis: Mining opinions, sentiments, and emotions: Cambridge University Press.

Liu, B., & Zhang, L. (2012). A survey of opinion mining and sentiment analysis Mining Text Data (pp. 415-463): Springer.

Mamdani, E. H. (1974). Application of fuzzy algorithms for control of simple dynamic plant. Paper presented at the Proceedings of the Institution of Electrical Engineers.

Pang, B., & Lee, L. (2008). Opinion mining and sentiment analysis. Foundations and trends in information retrieval, 2(1-2), 1-135.

Rana, T. A., & Cheah, Y. N. (2016). Aspect extraction in sentiment analysis: comparative analysis and survey. Artificial Intelligence Review, 1-25

Rosier, A., Burgun, A., & Mabo, P. (2008). Using regular expressions to extract information on pacemaker implantation procedures from clinical reports. Paper presented at the AMIA Annual Symposium Proceedings.

Ross, T. J. (2009). Fuzzy logic with engineering applications: John Wiley & Sons.

Rushdi Saleh, M., Martín-Valdivia, M. T., Montejo-Ráez, A., & Ureña-López, L. (2011). Experiments with SVM to classify opinions in different domains. Expert Systems with Applications, 38(12), 14799-14804.

Samuel, O., Omisore, M., & Ojokoh, B. (2013). A web based decision support system driven by fuzzy logic for the diagnosis of typhoid fever. Expert Systems with Applications, 40(10), 4164-4171.

Simpson, M. S., & Demner-Fushman, D. (2012). Biomedical text mining: A survey of recent progress Mining Text Data (pp. 465-517): Springer.

Swaminathan, R., Sharma, A., & Yang, H. (2010). Opinion mining for biomedical text data: Feature space design and feature selection. Paper presented at the The Ninth International Workshop on Data Mining in Bioinformatics, BIOKDD.

Turney, P. D. (2002). Thumbs up or thumbs down?: semantic orientation applied to unsupervised classification of reviews.

Paper presented at the Proceedings of the 40th annual meeting on association for computational linguistics.

Yalamanchi, D. (2011). Sideffective-system to mine patient reviews: sentiment analysis. Rutgers University-Graduate School-New Brunswick.

Yu, H., & Hatzivassiloglou, V. (2003). Towards answering opinion questions: Separating facts from opinions and identifying the polarity of opinion sentences. Paper presented at the Proceedings of the 2003 conference on Empirical methods in natural language processing.

Zadeh, L. A. (1965). Fuzzy sets. Information and control, 8(3), 338-353.

Zhang, L., & Liu, B. (2011). Identifying noun product features that imply opinions. Paper presented at the Proceedings of the 49th Annual Meeting of the Association for Computational Linguistics: Human Language Technologies: short papers-Volume 2.

Zhou, X., Han, H., Chankai, I., Prestrud, A., & Brooks, A. (2006). Approaches to text mining for clinical medical records. Paper presented at the Proceedings of the 2006 ACM symposium on Applied computing.

Zhou, X., Han, H., Chankai, I., Prestrud, A. A., & Brooks, A. D. (2005). Converting semi-structured clinical medical records into information and knowledge. Paper presented at the Data Engineering Workshops, 2005. 21st International Conference on.